\documentclass{article}

     \usepackage[final]{neurips_2021}

\usepackage[utf8]{inputenc} 
\usepackage[T1]{fontenc}    
\usepackage{hyperref}       
\usepackage{url}            
\usepackage{booktabs}       
\usepackage{amsfonts}       
\usepackage{nicefrac}       
\usepackage{microtype}      
\usepackage{xcolor}         
\usepackage{graphicx} 
\usepackage{multirow}
\usepackage{dsfont}
\usepackage{float}
\usepackage{natbib}
\setcitestyle{square,numbers}
\title{A 3D-Shape Similarity-based Contrastive Approach to Molecular Representation Learning}

\author{
Austin Atsango,$^{1}$ \quad Nathaniel Diamant,$^{2}$ \quad Ziqing Lu,$^2$ \quad \\ \textbf{Tommaso Biancalani,}$^2$ \quad \textbf{Gabriele Scalia,}$^{2\dagger}$ \quad \textbf{Kangway V. Chuang}$^{2\dagger}$\\
$^1$Department of Chemistry, Stanford University \\ $^2$Department of Artificial Intelligence and Machine Learning,\\ Genentech Research and Early Development\\
\texttt{$^\dagger$\{scalia.gabriele,chuang.kangway\}@gene.com}\\
}

\begin{document}

\maketitle

\begin{abstract}
Molecular shape and geometry dictate key biophysical recognition processes, yet many graph neural networks disregard 3D information for molecular property prediction. Here, we propose a new contrastive-learning procedure for graph neural networks, \textbf{Mol}ecular \textbf{C}ontrastive \textbf{L}e\textbf{a}rning from \textbf{S}hape \textbf{S}imilarity (\textbf{MolCLaSS}), that implicitly learns a three-dimensional representation. Rather than directly encoding or targeting three-dimensional poses, \textbf{MolCLaSS} matches a similarity objective based on Gaussian overlays to learn a meaningful representation of molecular shape. We demonstrate how this framework naturally captures key aspects of three-dimensionality that two-dimensional representations cannot and provides an inductive framework for scaffold hopping.

\end{abstract}

\section{Introduction and Background}

Molecular shape is critical for biophysical processes, yet encoding relevant three-dimensional features remains challenging for many molecular property prediction tasks, especially when an understanding of three-dimensional shape is limited or unknown \cite{Nicholls2010-rm}. Numerous methods have been developed to effectively encode individual conformers that are both appropriate and highly-effective for conformer-level prediction tasks such as predicting quantum chemical properties of single conformational poses \citep{Axen2017e3fp, Gilmer2017-nh, Schutt_undated-vi, Gasteiger_undated-za, Liu2022-by}. However, these approaches are poorly suited for representing complete molecules since relying on a single low-energy conformer to represent a diverse conformational ensemble is inherently limiting. \citet{Dietterich1997-bm} first recognized and addressed this challenge in the development of the multiple-instance learning framework \citep{Herrera2016-va}. Recent studies have explored deep multiple-instance learning approaches for learning on conformational ensembles \citep{Chuang2020-mr, Axelrod2020-ez}, yet are computationally-demanding due to the need to encode each conformer independently. Furthermore, methods that encode three-dimensional information often do not provide a strong performance benefit over 2D baselines~\citep{Axelrod2020-ez}.

Herein, we propose to inject graph neural networks with \textit{implicit} 3D shape information through a supervised contrastive approach. Learning implicit 3D representations have been recently explored by \citet{Stark2022-vm} and  \citet{liu2022pretraining}, but these approaches do not consider three-dimensional relationships between molecules and hence do not learn a direct measure of molecular shape similarity. In contrast, we propose to learn key features of molecular shape through direct comparison with the use of 3D molecular similarity kernels based on Gaussian overlays \cite{Grant1995-by}. Specifically, methods such as the Rapid Overlay of Chemical Structures (ROCS) \cite{Hawkins2007-lp} provide a fast and scalable method for matching molecular shapes based both on molecular volumes and electrostatic matching on predefined pharmacophore features. 

\begin{figure}[t]
    \centering
        \includegraphics[width=13.80cm]{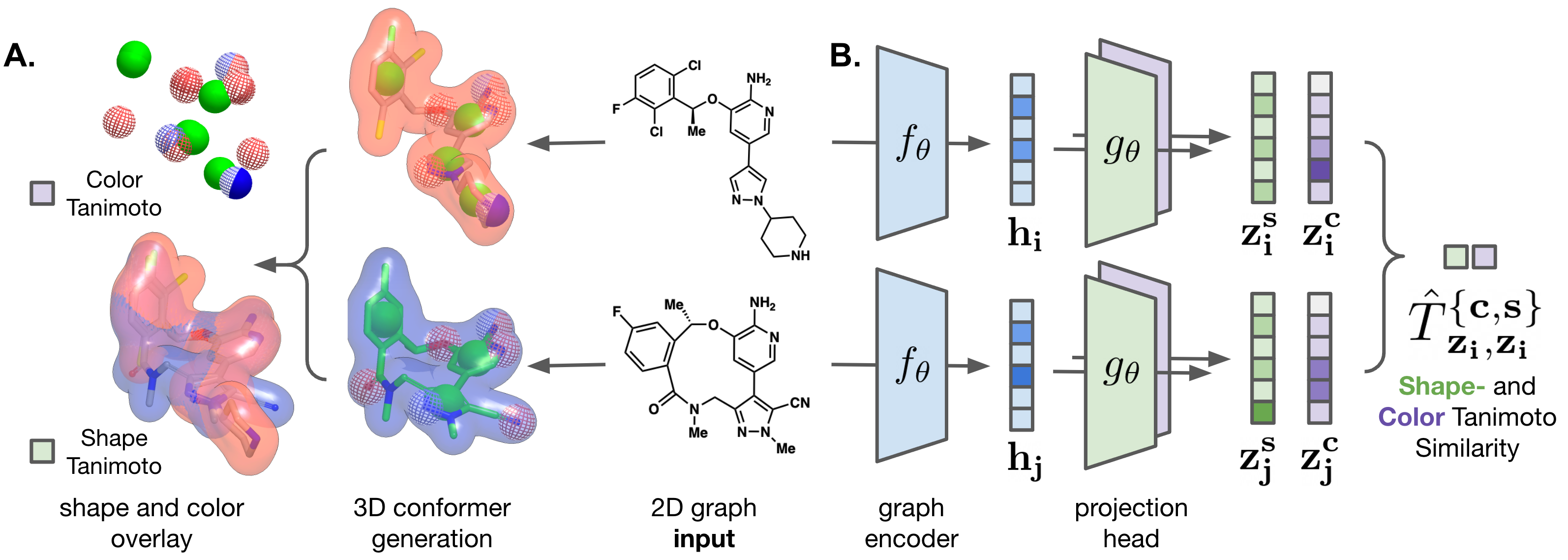}
    \caption{Our proposed model couples a pairwise, Gaussian shape- and pharmacophore similarity objective to a graph contrastive learning procedure to learn molecular embeddings with implicit 3D-information. {\bf A.} For each input molecule, we sample a conformational ensemble and perform OpenEyeROCS to find a maximum overlay via ShapeTanimoto and ColorTanimoto similarities. {\bf B.} The resulting pairwise, 3D shape similarities become the objective of our graph contrastive learning procedure on 2D graphs \textit{only}, which are approximated using a Tanimoto kernel function.}
    \label{fig:figure_1}
    \vskip -0.1in
\end{figure}

Our framework, \textbf{Mol}ecular \textbf{C}ontrastive \textbf{L}e\textbf{a}rning on \textbf{S}hape \textbf{S}imilarity (MolCLaSS), naturally aligns a pairwise shape similarity objective with supervised contrastive learning framework to learn meaningful representations of molecular shape (Figure \ref{fig:figure_1}). Our work contributes directly in two ways: 1) MolCLaSS provides fast, scalable, and inductive approximation of 3D-shape similarity scores between molecules directly from their 2D graphs, and does not require conformer generation and shape alignment for inference, and 2) we demonstrate how MolCLaSS learns meaningful molecular embeddings that naturally capture 3D-shape and features that 2D, topological methods cannot. The resulting pre-trained encoder can be used for downstream molecular shape tasks.

\section{Related Work}
\label{related_work}
\paragraph{Graph Neural Networks for Small Molecules} Graph neural networks have been widely developed for predicting small molecule properties and activities \citep{Duvenaud2015-wo, Kearnes2016-ab, Yang2019-he} on both 2D and 3D tasks. The properties of individual conformers have been effectively modeled by leveraging 3D spatial features for a range of quantum chemical and property prediction tasks \citep{Gilmer2017-nh, Schutt_undated-vi, Gasteiger_undated-za, Liu2022-by}. \citet{Adams2021-vt} recently described a hybrid approach for conformationally-invariant 3D approaches. Furthermore, "4D" multiple-instance learning methods have recently been developed that operate over sets of conformers that are each modeled as a graph \citep{Chuang2020-mr, Axelrod2020-ez}. Our work builds on this prior work to learn improved and compact representations. 

\paragraph{Pre-Training Graph Neural Networks} Graph pretraining methods are an active area of research. \citet{Hu2020Strategies} first reported a pretraining strategy based on self-supervised node pre-training, followed by supervised pre-training with masking. \citet{You2020GraphCL} recently demonstrated the effectiveness of a contrastive learning approach on graphs, which was further extended by \citet{Wang2022-bg, Wang2022-fx}. Recently, \citet{Stark2022-vm} and \citet{liu2022pretraining} developed self-supervised approaches to maximize three-dimensional information for 2D-graph neural network pretraining with promising results. These approaches aim to maximize mutual information with the goal of matching a 2D representation to a three-dimensional pose, but they do not explicitly consider three-dimensional similarity between molecules. Our work here presents a supervised contrastive approach for pretraining neural networks that complements the self-supervised approaches above.

\paragraph{Kernel-Based Approximations of Molecular Shape} Our work is closely related to prior work on learning low dimensional embeddings from molecular similarity kernels. \citet{Raghavendra2007-en} introduced a method to learn molecular basis vectors by directly decomposing Tanimoto similarities. The SCISSORS method by Haque \& Pande \citep{Haque2010-he, Haque2011-vw, Kearnes2014-cv} generalizes a kernel PCA approach to molecular similarity measures, including ROCS, that provides a fast approximation for molecular similarity. These prior approaches are naturally transductive; to obtain new scores, molecules must be scored against the resulting basis set and predicted based on a least-squares estimate. Our approach couples the intuition of \citet{Haque2010-he} with modern graph neural networks to directly learn a meaningful molecular space that is naturally inductive, i.e. can generalize to new and unseen molecules without the need for additional conformer generation and scoring.

\section{Problem Formulation and Methods}
We invoke a variation of the classic similar property principle and assume that two molecules are similar if they can adopt similar molecular shapes \cite{Johnson1990-uz, Todeschini2008-ws}. Prior approaches for encoding three-dimensional information typically operate on a molecular input $x_i$ and corresponding spatial features $s_i$ that includes explicit atomic coordinate, distance, or angle information. These approaches rely on designing an expressive transformation $f_\theta(x_i, s_i)$ that can accurately learn to map similar molecules to similar parts of chemical space, and assume that a single or several low-energy conformers encode relevant geometry. Rather than operate over explicit spatial representations, we instead adopt an \textit{implicit} strategy that leverages a predefined similarity kernel over pairs of molecular inputs $k(x_i, x_j)$. Critically, by leveraging a well-defined 3D similarity function using inner products, we avoid the need to explicitly encode spatial features $s_i$, and can  focus on learning an invariant function $f_\theta(x_i)$. 

As illustrated in Figure \ref{fig:figure_1}, our approach naturally fits a supervised contrastive learning framework: given a set of molecules, our goal is to learn an expressive representation satisfies a pairwise similarity constraint $k(x_i, x_j)$. Here, we decompose this representation into  graph encoder $f_\theta$ and projection heads $g_\theta$ to flexibly model multiple outputs. In this study, we use Gaussian shape and color overlays as an intuitive measure of shape similarity \citep{Grant1995-by, Nicholls2010-rm}, and follow the insight of \citet{Haque2010-he, Haque2011-vw} to leverage the Tanimoto kernel \citep{Ralaivola2005-rj}:

\begin{equation} \label{eq:tanimoto-kernel}
T(\mathbf{z}_i, \mathbf{z}_j) = \frac{
\mathbf{z}_i \cdot \mathbf{z}_j}{
\mathbf{z}_i \cdot \mathbf{z}_i+ 
\mathbf{z}_j \cdot \mathbf{z}_j - 
\mathbf{z}_i \cdot \mathbf{z}_j}
\end{equation}

This interpretation defines a molecular embedding space, where three-dimensional shape similarities can be conveniently modeled based on inner products. Rather than approximate embeddings $\mathbf{z_i}$ via linear decomposition methods, our objective is to learn an inductive model that generates $\mathbf{z_i}$ directly from a molecular graph. Here, we use an encoder based on the Graph Isomorphism Network with Edge features \cite{Hu2020Strategies} followed by two projection heads to model Shape and Color separately:

\begin{equation}
\mathbf{v}^{t+1}_p = q_{\mathbf{\theta}} ( (1 + \epsilon) \cdot \mathbf{v}^{t}_p +\sum_{q \in \mathcal{N}(p)} \sigma
        ( \mathbf{v}^{t}_q + \mathbf{e}_{p,q} ))  
\end{equation}
\begin{equation}
        \hspace{0pt} \textrm{with} \hspace{10 pt} f_\theta(\mathbf{x}_i) = \mathbf{h}_i = \sum_{p \in G} \mathbf{v}^T_p
\end{equation}

\begin{equation}
        \mathbf{z}_i^{\{c, s\}} = g_\theta(\mathbf{h}_i) = U^{\{c, s\}}\sigma(V^{\{c, s\}}\mathbf{h}_i)
\end{equation}

Above, $\mathbf{v}^t_p$ corresponds to the hidden state of the node $p$ at step $t$ (final step $T$), with $\mathbf{h}_i$ as the final graph representation of molecule $i$ using sum pooling. The projection heads $g_\theta(\mathbf{h}_i)$ are parameterized by Color- and Shape-dependent MLPs with trainable weight matrices $U$ and $V$, and $\sigma$ is a  ReLU nonlinearity. Finally, we optimize the network to directly predict ShapeTanimoto and ColorTanimoto scores via minimization of the following loss function:

\begin{equation}
\mathcal{L}_{\{s,c\}} = \frac{1}{N^2}\sum_{i=1}^{N} \sum_{j=1}^{N} \left( T_{\{s,c\}}({\mathbf{z}_i, \mathbf{z}_j}) - k_{\{s,c\}}({\mathbf{x}_i, \mathbf{x}_j})\right)^2 \ 
\hspace{5 pt} \textrm{with} \hspace{5 pt} \mathcal{L}= \mathcal{L}_{s} + \lambda \mathcal{L}_{c}
\end{equation}

Here, we define the loss as the mean squared error over all the pairs in a batch (size $N$) between the predicted Color and Shape scores based on the Tanimoto kernel (Eq. \ref{eq:tanimoto-kernel}) and the calculated Gaussian overlay $k_{\{s,c\}}\left({\mathbf{x}_i, \mathbf{x}_j}\right)$ based on conformer generation and alignment. We balance the individual objectives ${L}_{s}$ and ${L}_{c}$ with the tuneable hyperparameter $\lambda$ to adjust the influence of pharmacophore features (but typically set to~$1$).

\section{Experiments and Results}
\label{Experiments and Results}

\begin{table}[t]
  \centering
  \label{table1}
  \begin{tabular}{cccccccccc}
    \multirow{2}{*}{}&
    \multicolumn{1}{c}{}&
      \multicolumn{3}{c}{\bf ShapeTanimoto} &
      \multicolumn{3}{c}{{\bf ColorTanimoto}} \\
    \toprule
    Model & $n$ & $r$ & $R^{2}$  & MAE & $r$ & $R^{2}$   & MAE \\
    \midrule
    2D Tanimoto & – & 0.000  & –  & – & 0.000 & – & – \\
    
    ECFP4 + FF-NN & 10k  & 0.733  & 0.513 & 0.0473 & 0.502 & 0.144 & 0.0484 \\
    ECFP4 + FF-NN & 50k  & 0.793  & 0.610 & 0.0419 & 0.593 & 0.243 & 0.0456 \\
    ECFP4 + FF-NN & 100k  & 0.822  & 0.660 & 0.0388 & 0.636 & 0.327 & 0.0428\\

    MolCLaSS  & 10k  & 0.818 & 0.653 & 0.0396 & 0.608 & 0.295 & 0.0442 \\
    MolCLaSS   & 50k  & 0.876  & 0.757 & 0.0327 & 0.718 & 0.484 & 0.0374\\
    \textbf{MolCLaSS}   & \textbf{100k} & \textbf{0.893} & \textbf{0.794}  & \textbf{0.0301} & \textbf{0.748} & \textbf{0.537} & \textbf{0.0353} \\

    \bottomrule
  \end{tabular}
  \vspace{0.2 in}
  \caption{Model performance for prediction of ShapeTanimoto and ColorTanimoto scores from 2D graphs. All results are reported on a random, independent test set of 49,621 molecules corresponding to 1.23 billion pairwise similarity scores. For each training set size $n$, we report Pearson's correlation coefficient $r$, the coefficient of determination $R^2$, and the mean absolute error (MAE) for all 1.23 billion pairwise scores.}
\end{table}

\paragraph{Prediction of Shape- and Color-Tanimoto Scores from 2D Graphs} We systematically investigated the ability of contrastive models to directly predict ShapeTanimoto and ColorTanimoto scores of drug-like molecules from the ChEMBL database \citep{Gaulton2012-db, Mendez2019-az}. For these studies, we generated a complete all-by-all similarity matrix of Shape- and ColorTanimoto scores \citep{Grant1995-by, Hawkins2007-lp} for 100k molecules from ChEMBL, with maximum overlay scores recorded from pairwise comparisons of up to 10 conformers generated per molecule \citep{Hawkins2010-xq, Hawkins2012-aj} (see Appendix for complete details). We refer to the complete data set as \texttt{ROCS100k}. We assessed the ability of both fingerprint- and graph-based models to accurately predict pairwise, 3D similarity scores directly from their 2D graph representations. Although thissetup requires computationally-intensive conformer generation and exhaustive similarity scoring, this cost is amortized over the training data. At inference time, predicted Shape- and ColorTanimoto similarities are directly obtained and circumvent the need for explicit three-dimensional representations.

As illustrated in Table \ref{table1}, we directly compared our approach to a Morgan fingerprint \citep{Rogers2010-mp} and dense neural network baseline, with positive performance gains for increasingly large data set sizes. Our graph neural network-based approach based on GINEConv graph layers with independent projection heads for ShapeTanimoto and ColorTanimoto predictions exhibits a clear improvement over hashed fingerprint representations, even with significantly less data. For example, MolCLaSS trained with only 10k examples achieves nearly identical performance to fingerprint-based models trained on 100k examples. Critically, these dense networks learn a non-trivial transformation of the input data. Indeed, as shown in Table \ref{table1} (first row), there is nearly no correlation between bulk Tanimoto scores on 2D representations and their 3D Tanimoto scores.

Notably, the MolClASS network can directly predict 3D similarity scores with good accuracy (ShapeTanimoto MAE$ = 0.030$, ShapeTanimoto MAE$ = 0.035$) at only a fraction of the computational cost. At inference, predicting 3D similarity scores on tens of thousands of molecules takes only seconds, replacing both the need for conformer generation and overlays and representing an improvement of nearly $10^4 - 10^5 $ times in speed.

\begin{figure}[t]
    \centering
    \includegraphics[width=13.8cm]{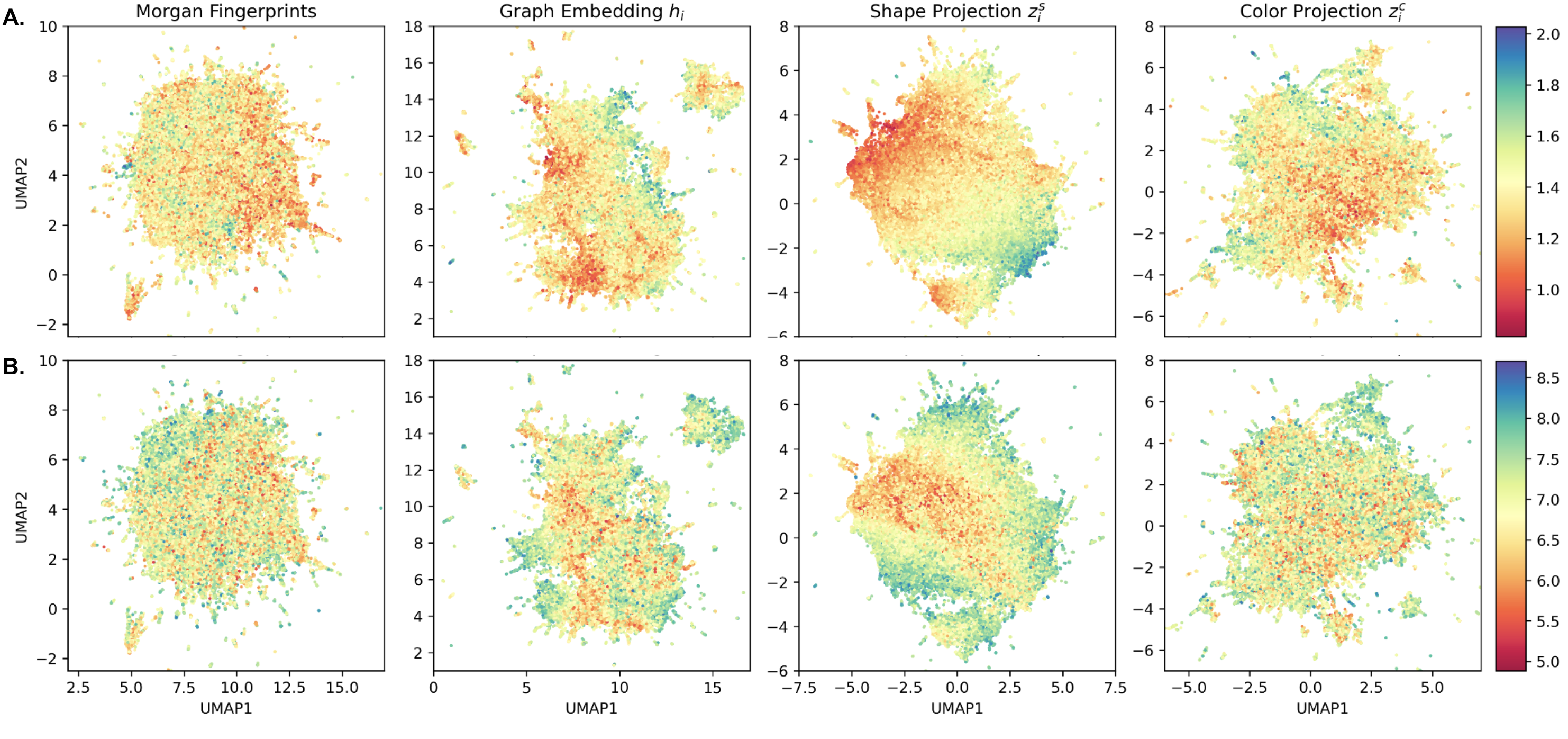}
    \vskip -0.1 in
    \caption{Comparison of MolClaSS latent graph encodings and projection heads to topological fingerprints reveals that MolCLaSS provides meaningful latent organization by three-dimensional features. Latent representations colored by the \textbf{A. radius of gyration (log-scale)} and \textbf{B. first principal moment of inertia (log-scale) } for a low-energy conformer. The graph embeddings $\mathbf{h}_i$ and shape projection $\mathbf{z}^s_j$ learn more localized structure corresponding to human interpretable features.}
    \label{fig:umap}
\end{figure}

\paragraph{MolCLaSS Representations Capture 3D Shape Similarity.}
Given the strong performance of MolCLaSS, we investigated the qualities of the learned molecular embeddings. We specifically visualized our hold-out test set of ChEMBL molecules in their graph embedding ($\mathbf{h_i}$) and projection heads ($\mathbf{z^s_i, z_i^c}$) (Figure~\ref{fig:umap}) against a fingerprint baseline, and colored them based on calculated three-dimensional descriptors \cite{Todeschini2008-ws, Landrum2006-si} of their low-energy conformers, including radius of gyration (Figure~\ref{fig:umap}A) and the first principal moment of inertia (Figure~\ref{fig:umap}B). As shown, both the graph-layers and shape-projection layer provide clear localization based on these 3D properties when compared to Morgan fingerprints. As expected, the color projection head does not exhibit the same localization, as pharmacophore features are less dependent on overall molecular shape.

\begin{figure}[h]
    \centering
    \includegraphics[width=13.8cm]{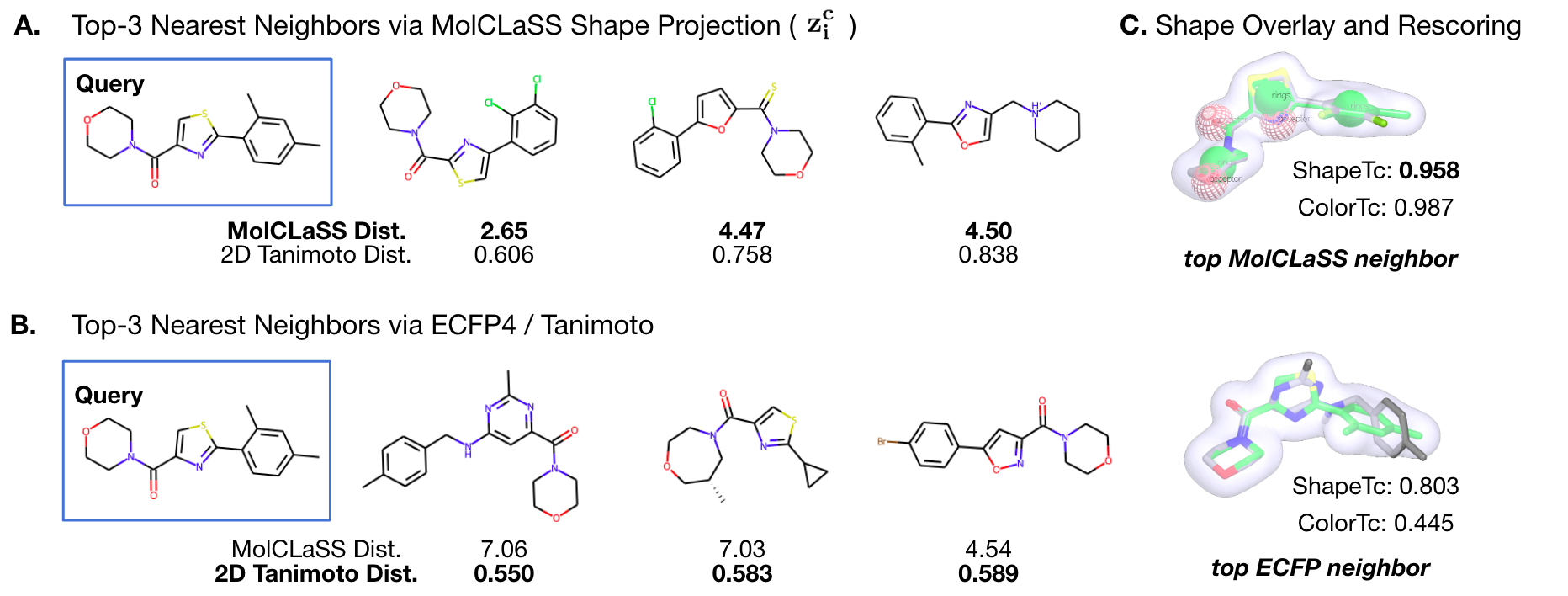}
    \caption{Nearest neighbors analysis using learned latent representations. We analyze the test set using \emph{k}-nearest neighbors to retrieve similar hits based on a given query (top right). \textbf{A.} Nearest neighbors with MolCLaSS (Euclidean distance) preserves scaffold shape and can hop between scaffolds through core mutations (top row). \textbf{B.} The same analysis performed with ECFP4 fingerprints and Tanimoto distance (bottom row). Topological fingerprints heavily emphasize subgraph matches. Although diverse hits are found, nearest neighbors have a lower shape similarity. \textbf{C.} Reanalysis of top hits by fastROCS demonstrates how MolCLaSS accurately identifies close three-dimensional matches (top right), whereas 2D fingerprints produce lower-quality matches.}
    \label{fig:nn}
\end{figure}

A key consequence of the Tanimoto similarity objective (Eq. \ref{eq:tanimoto-kernel}) is that it induces a Euclidean structure over the projected vector space $z$ \cite{McDermott2021-xh}. Indeed, we found an excellent correlation (Pearson $r = 0.87$) between pairwise Euclidean distance and ShapeTanimoto scores (see Appendix). To further probe the difference in representations we performed a nearest neighbors analysis using the shape projections in MolCLaSS (Figure \ref{fig:nn}). In our analysis, we consistently find that MolCLaSS preserves molecular shape and size. Scaffolds hops based on ring mutations are found nearby while maintaining excellent overall shape similarity. In contrast, topological fingerprints largely favor substructure matching and exhibits a wider range of molecular shapes. Together, these two studies illustrate how the MolCLaSS framework can capture relevant shape measures through a supervised approach.

\section{Conclusions and Future Directions}
Our studies above outline a preliminary roadmap for learning three-dimensional representations based on 3D similarities. In contrast to prior work, MolCLaSS learns via supervised pairwise comparisons, and hence is able to relate and differentiate molecules of varying size and shape. We have demonstrated how the MolCLaSS network itself provides a direct and fast approximation method for approximating shape- and 3D pharmacophore-based, and further illustrated how meaningful three-dimensional features are naturally learned through this inductive framework. Our ongoing work seeks to further improve the MolCLaSS framework to improve predictive performance and to explore its application as a pretrained model for broad molecular property prediction tasks.

\begin{ack}
We gratefully acknowledge members of the Department of Artificial Intelligence and Machine Learning at Genentech Research and Early Development for helpful feedback and support. 
\end{ack}

\bibliography{bibliography}

\newpage
\section*{Appendix}\label{Appendix}

All experiments were performed using Python and standard numerical libraries. For cheminformatics analysis, all molecules were processed using either OpenEye Applications and Toolkits or the open-source cheminformatics library RDKit \citep{Landrum2006-si}. We implemented all experiments in Python using PyTorch \texttt{1.11} \citep{Paszke2019-fn} and PyTorch Geometric \citep{Fey2019-ki}.

\subsection*{Appendix A: Molecular Dataset Properties and Creation} 
\begin{figure}[h]
    \centering
    \includegraphics[width=13.8cm]{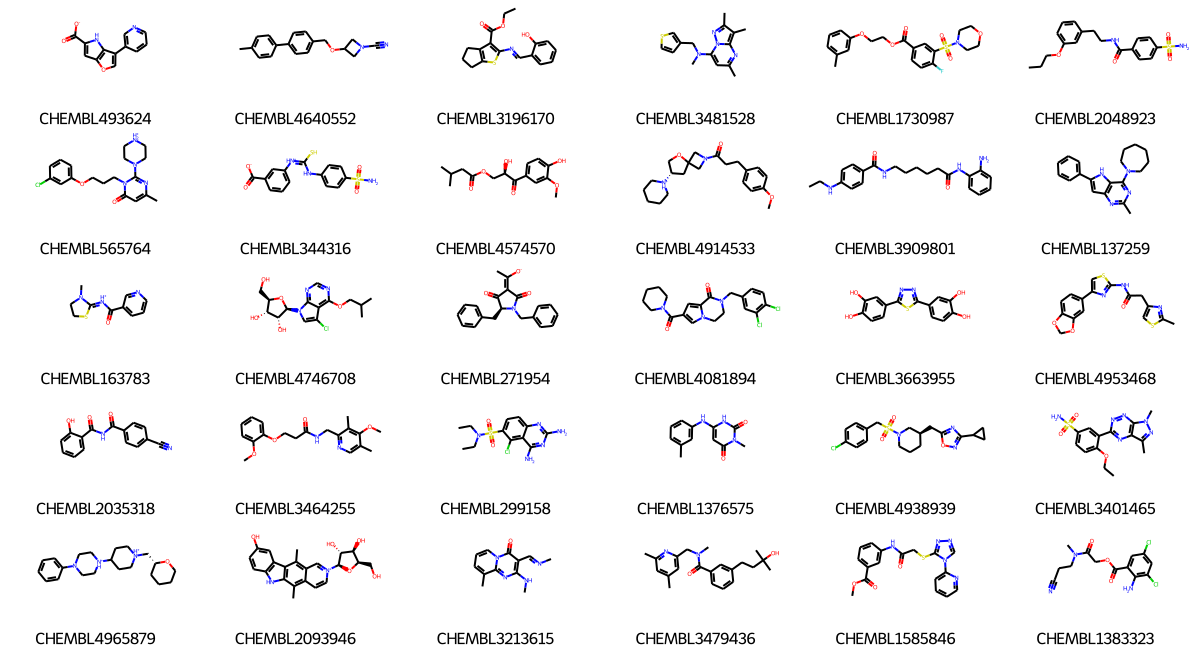}
    \caption{Random selection of 50 examples from ChEMBL31 for generating the ROCS100k dataset.}
    \label{fig:molsamples}
\end{figure}

All data were downloaded directly from ChEMBL31 release of the ChEMBL database \citep{Gaulton2012-db, Mendez2019-az}. Molecules were filtered based on OpenEye Filter using the drug-likeness that limits. The resulting detailed statistics of the molecules and some of their properties are shown in Table \ref{trainmolprops}, with a representative sample of molecules shown in Figure \ref{fig:molsamples}.

\subsubsection*{OMEGA Conformer Generation}
For each molecule, we generate a conformational ensemble using OpenEye Applications (\texttt{2022.1.1}) using OMEGA (\texttt{v.4.2.0}) \citep{Hawkins2010-xq, Hawkins2012-aj}. Conformational ensembles were generated with the optimized default \texttt{fastROCS} settings with multiprocessing: \texttt{maxconfs=10, ewindow=15, flipper=False, mpi=128} which been shown to accurately recapitulate binding poses. Molecules with ambiguous or undefined stereochemistry were dropped during the conformer generation process. 

\subsubsection*{OpenEye fastROCS Scoring}

We use OpenEye's GPU-accelerated fastROCS Toolkit (\texttt{v.2.2.2.1}) to calculate ShapeTanimoto and ColorTanimoto scores. For each conformer database, we generate an all-by-all similarity matrix that scores every conformer of every molecule against all conformers of the rest of the database, saving the maximum score between two molecules. In practice, generation of the ROCS100k training dataset generates 4.95 billion ShapeTanimoto and ColorTanimoto scores, each, and requires 7 hours on 3 NVIDIA 3090 RTX GPUs with 80 processes. 

\paragraph*{Computational Complexity of All-by-All Conformer ROCS} Exhaustive similarity comparisons between $n$ molecules scales at $\mathcal{O}(n^2)$. As an upper limit, for $k$ conformers are generated per molecule, generation of the all-conformer by all-conformer similarity matrix scales at $\mathcal{O}(k^2n^2)$. For symmetric similarity measures like ColorTanimoto and ShapeTanimoto, $\mathcal{O}((n)(n-1)/{2})$ comparisons are required with no self-comparisons. Similarity, for $k$ sampled conformers we require $\mathcal{O}((kn)(kn-1)/{2})$ pairwise comparisons. 


\begin{table}[h]
  \caption{Detailed Statistics of the ROCS100k Training Dataset ($n$ = 100,000).}
  \label{trainmolprops}
  \centering
  \begin{tabular}{cccccc}
    \toprule
    Property & min.  & max. & mean & median & std.\\
    \midrule
    Sampled Conformers ($k$) & 1 & 10 & 9.48 & 10 & 1.78 \\
    Heavy Atom Count & 15 & 35 & 23.88 & 24 & 4.71 \\
    Molecular Weight & 220.20 & 598.03 & 337.73 & 334.42 & 66.38 \\
    Rotatable Bonds & 0  & 11 & 4.51 & 4 & 1.99  \\
    Aromatic Rings & 0 & 12 & 3.14 & 3 & 1.01 \\
    H-Bond Donors & 0 & 5 & 1.44 & 1 & 0.98 \\
    H-Bond Acceptors & 0 & 12 & 4.34 & 4 & 1.72 \\
    Heteroatoms & 2 & 14 & 6.17 & 6 & 1.84 \\
    \bottomrule
  \end{tabular}
\end{table}

\begin{table}[h]
  \caption{Detailed Statistics of the ROCS50k Test Dataset ($n$ = 49,621).}
  \label{testmolprops}
  \centering
  \begin{tabular}{cccccc}
    \toprule
    Property & min.  & max. & mean & median & std.\\
    \midrule
    Sampled Conformers ($k$) & 1 & 10 & 9.53 & 10 & 1.70 \\
    Heavy Atom Count & 15 & 35 & 23.66 & 23 & 4.77 \\
    Molecular Weight & 220.17 & 557.52 & 335.4 & 331.42 & 67.56 \\
    Rotatable Bonds & 0  & 10 & 4.42 & 4 & 1.96  \\
    Aromatic Rings & 0 & 8 & 3.08 & 3 & 1.00 \\
    H-Bond Donors & 0 & 5 & 1.31 & 1 & 0.97 \\
    H-Bond Acceptors & 0 & 12 & 4.62 & 5 & 1.70 \\
    Heteroatoms & 2 & 14 & 6.37 & 6 & 1.85 \\
    \bottomrule
  \end{tabular}
\end{table}

\subsection*{Appendix B: Network Architecture and Training Details}

All studies were trained using the generated ROCS100k dataset. The size dependence studies illustrated in \ref{table1} using 10k and 50k examples use subsets of the full 100k molecules. A separate validation set of 8,548 molecules from ChEMBL31 were additionally used for model tuning and selection. All results in Table \ref{table1} are reported on an independent, random test set of 49,621 molecules from ChEMBL31.

All neural networks were trained  using the Adam optimizer (learning rate = $1 \times 10^{-3}$ to $1 \times 10^{-4}$) and a batch size of 2048 for up to 4,000 epochs, using the early stopping criterion based on the validation set described above. The model architecture and hidden dimensions are specified in Appendix Table \ref{neural network}. All networks use five graph encoding layers with two project heads (single-hidden layer MLPs). The entire network is trained on Shape- and Color-Tanimoto targets bounded within [0,1], using mean squared error as the loss criterion and trained to early stopping.

\subsubsection*{Neural Network Performance}

\begin{figure}[h]
    \centering
    \includegraphics[width=13.8cm]{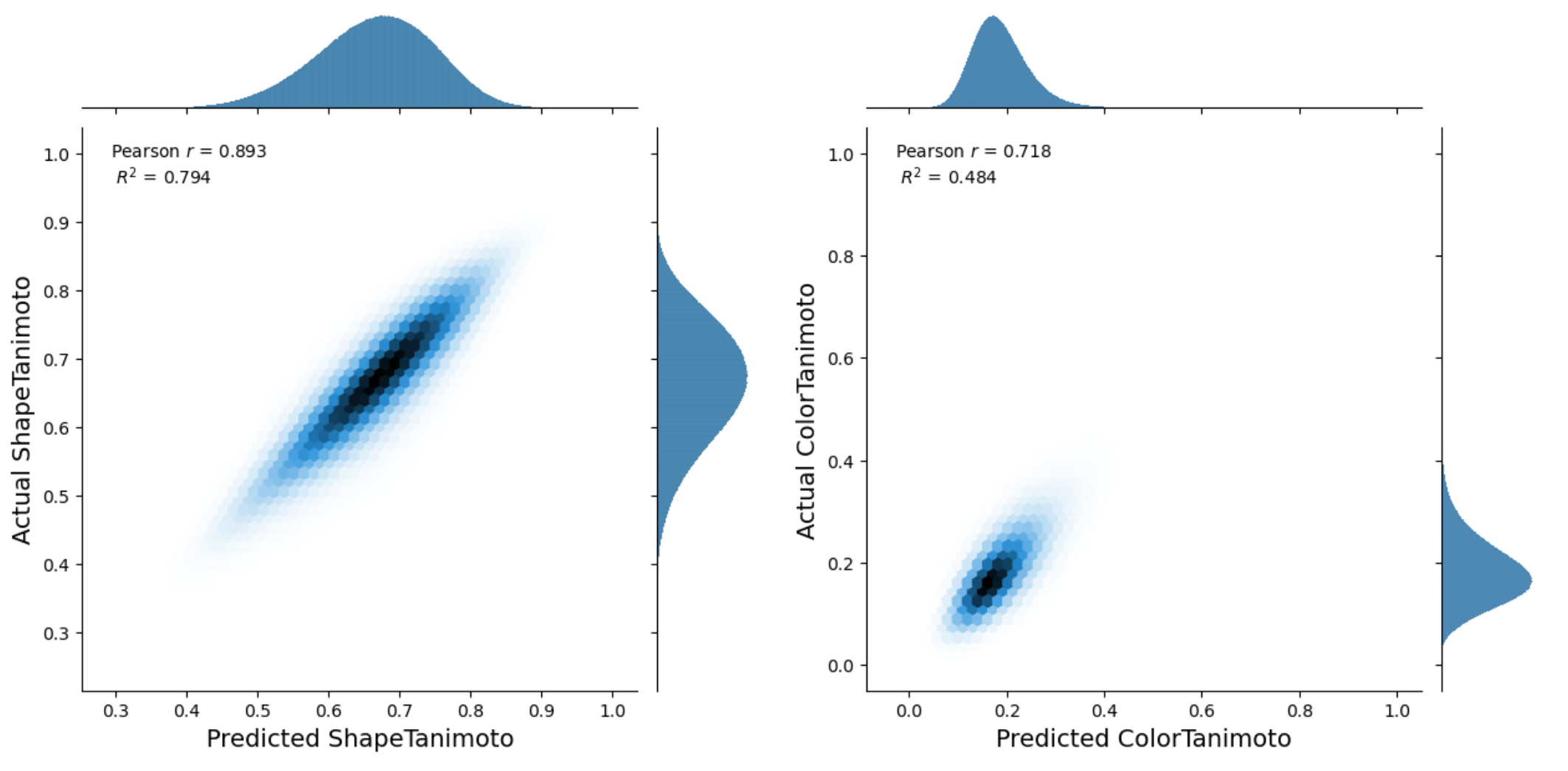}
    \caption{Performance of MolCLaSS for predicting ShapeTanimoto and ColorTanimoto trained on the full \texttt{ROCS100k} data set, corresponding to the final entry in Table \ref{table1}. results plotted are on the full 49,621 test set molecules.}
    \label{fig:nn_performance}
\end{figure}

Our summary of overall performance results and metrics for different model architectures are shown in Table \ref{table1}. Below, we include scatter plots of our best model trained on the full ROCS100k dataset in Figure \ref{fig:nn_performance}.

\subsubsection*{Correlation of Euclidean Distance with ShapeTanimoto and ColorTanimoto}

\begin{figure}[h]
    \centering
    \includegraphics[width=13.8cm]{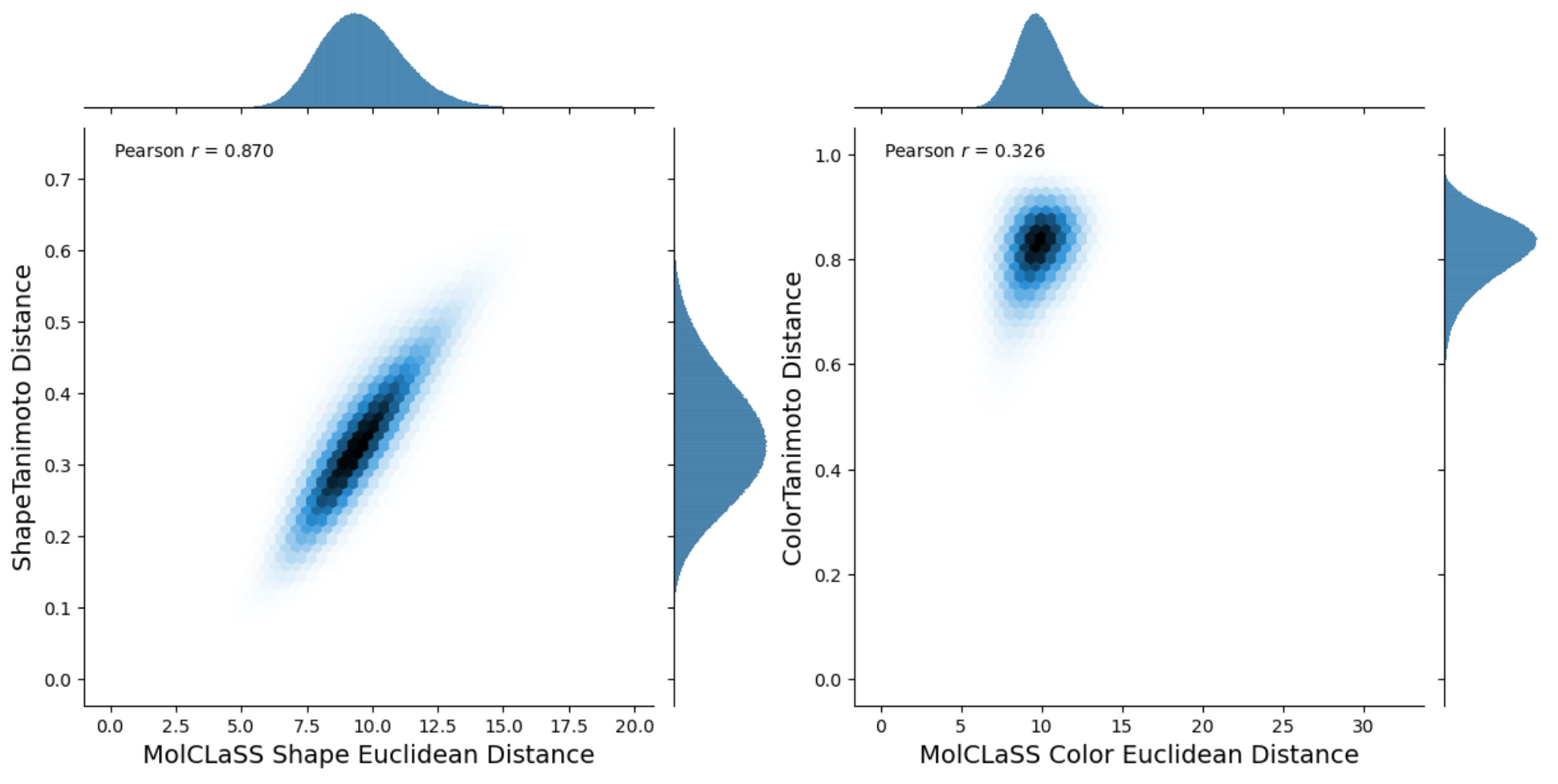}
    \caption{Plots comparing distance in ShapeTanimoto an ColorTanimoto, vs Euclidean distance for the Shape- and Color-projection heads. MolClaSS learns a meaningful structural embedding space where similar shapes are closer together in their latent representation (left), with a more modest correlation observed for ColorTanimoto (right). Correlations plotted for the unseen 49,621 molecules in the test set.} 
    \label{fig:euclideanperformance}
\end{figure}

The project heads $g_\theta$ ultimately learn a meaningful Euclidean distance. As shown in the plots below, molecules closer in the embedded latent space also tend to have a much higher Shape and ColorTanimoto score (Figure \ref{fig:euclideanperformance}).

\begin{table}[t]
  \caption{Neural Network Model Architectures}
  \label{neural network}
  \centering
  \begin{tabular}{lll}
    \toprule
    Model \& Module & Layer \& Description & $h_{dim}$ Sizes \\
    \midrule
    \textbf{MolCLaSS} \\
    Graph Encoder $f_\theta$ &  5 x [GINEConv + BatchNorm w/ ReLU]  &$ 5 \times (512 \rightarrow 1024 \rightarrow 512)$ \\
    Projection Head $g_\theta$ & MLP + ReLU  & $512 \rightarrow 1024 \rightarrow 256 $ \\
    \textbf{ECFP4 + FF-NN} \\
    Encoder $f_\theta$ & MLP w/ ReLU   & $2048 \rightarrow 2048 \rightarrow 512$\\
    Projection Head $g_\theta$ & MLP + ReLU  & $512 \rightarrow 1024 \rightarrow 256 $ \\
    \bottomrule
  \end{tabular}
\end{table}
\end{document}